\documentclass{article}

\usepackage{arxiv}

\usepackage[utf8]{inputenc} 
\usepackage[T1]{fontenc}    
\usepackage{hyperref}       
\usepackage{url}            
\usepackage{booktabs}       
\usepackage{amsfonts}       
\usepackage{nicefrac}       
\usepackage{microtype}      
\usepackage{lipsum}
\usepackage{graphicx}
\graphicspath{ {./images/} }

\usepackage{color}
\usepackage{lscape}
\usepackage{longtable}
\usepackage{geometry}
\usepackage{amsmath}
\usepackage{float}

\title{Analysis of the Evolution of Advanced Transformer-Based Language Models: Experiments on Opinion Mining}

\author{Nour Eddine Zekaoui \thanks{Corresponding author.} \hspace{1.0cm} Siham Yousfi \hspace{1.0cm} Maryem Rhanoui \hspace{1.0cm} Mounia Mikram \\ \\
  Meridian Team, LYRICA Laboratory\\
  School of Information Sciences\\
  Rabat, Morocco \\ \\
  \texttt{(nour-eddine.zekaoui, syousfi, mrhanoui, mmikram)@esi.ac.ma} \\
}

\begin{document}
\maketitle
\begin{abstract}
Opinion mining, also known as sentiment analysis, is a subfield of natural language processing (NLP) that focuses on identifying and extracting subjective information in textual material. This can include determining the overall sentiment of a piece of text (e.g., positive or negative), as well as identifying specific emotions or opinions expressed in the text, that involves the use of advanced machine and deep learning techniques. Recently, transformer-based language models make this task of human emotion analysis intuitive, thanks to the attention mechanism and parallel computation. These advantages make such models very powerful on linguistic tasks, unlike recurrent neural networks that spend a lot of time on sequential processing, making them prone to fail when it comes to processing long text. The scope of our paper aims to study the behaviour of the cutting-edge Transformer-based language models on opinion mining and provide a high-level comparison between them to highlight their key particularities. Additionally, our comparative study shows leads and paves the way for production engineers regarding the approach to focus on and is useful for researchers as it provides guidelines for future research subjects.
\end{abstract}


\section{Introduction}
Over the past few years, interest in natural language processing (NLP) \cite{chowdhary2020natural} has increased significantly. Today, several applications are investing massively in this new technology, such as extending recommender systems \cite{rhanoui2020hybrid}, \cite{trabelsi2021hybrid}, uncovering new insights in the health industry \cite{pandey2021comprehensive}, \cite{harnoune2021bert}, and unraveling e-reputation and opinion mining \cite{medhat2014sentiment}, \cite{sun2017review}. Opinion mining is an approach to computational linguistics and NLP that automatically identifies the emotional tone, sentiment, or thoughts behind a body of text. As a result, it plays a vital role in driving business decisions in many industries. However, seeking customer satisfaction is costly expensive. Indeed, mining user feedback regarding the products offered, is the most accurate way to adapt strategies and future business plans. In recent years, opinion mining has seen considerable progress, with applications in social media and review websites. Recommendation may be staff-oriented \cite{rhanoui2020hybrid} or user-oriented \cite{yousfi2017mixed} and should be tailored to meet customer needs and behaviors.

Nowadays, analyzing people’s emotions has become more intuitive thanks to the availability of many large pre-trained language models such as bidirectional encoder representations from transformers (BERT) \cite{devlin2018bert} and its variants. These models use the seminal transformer architecture \cite{vaswani2017attention}, which is based solely on attention mechanisms, to build robust language models for a variety of semantic tasks, including text classification. Moreover, there has been a surge in opinion mining text datasets, specifically designed to challenge NLP models and enhance their performance. These datasets are aimed at enabling models to imitate or even exceed human level performance, while introducing more complex features.

Even though many papers have addressed NLP topics for opinion mining using high-performance deep learning models, it is still challenging to determine their performance concretely and accurately due to variations in technical environments and datasets. Therefore, to address these issues, our paper aims to study the behaviour of the cutting-edge transformer-based models on textual material and reveal their differences. Although, it focuses on applying both transformer encoders and decoders, such as BERT \cite{devlin2018bert} and generative pre-trained transformer (GPT) \cite{radford2018improving}, respectively, and their improvements on a benchmark dataset. This enable a credible assessment of their performance and understanding their advantages, allowing subject matter experts to clearly rank the models. Furthermore, through ablations, we show the impact of configuration choices on the final results.

\section{Background}
\label{}
\subsection{Transformer}
\begin{figure}[ht]
    \centering
    \includegraphics[scale=0.60]{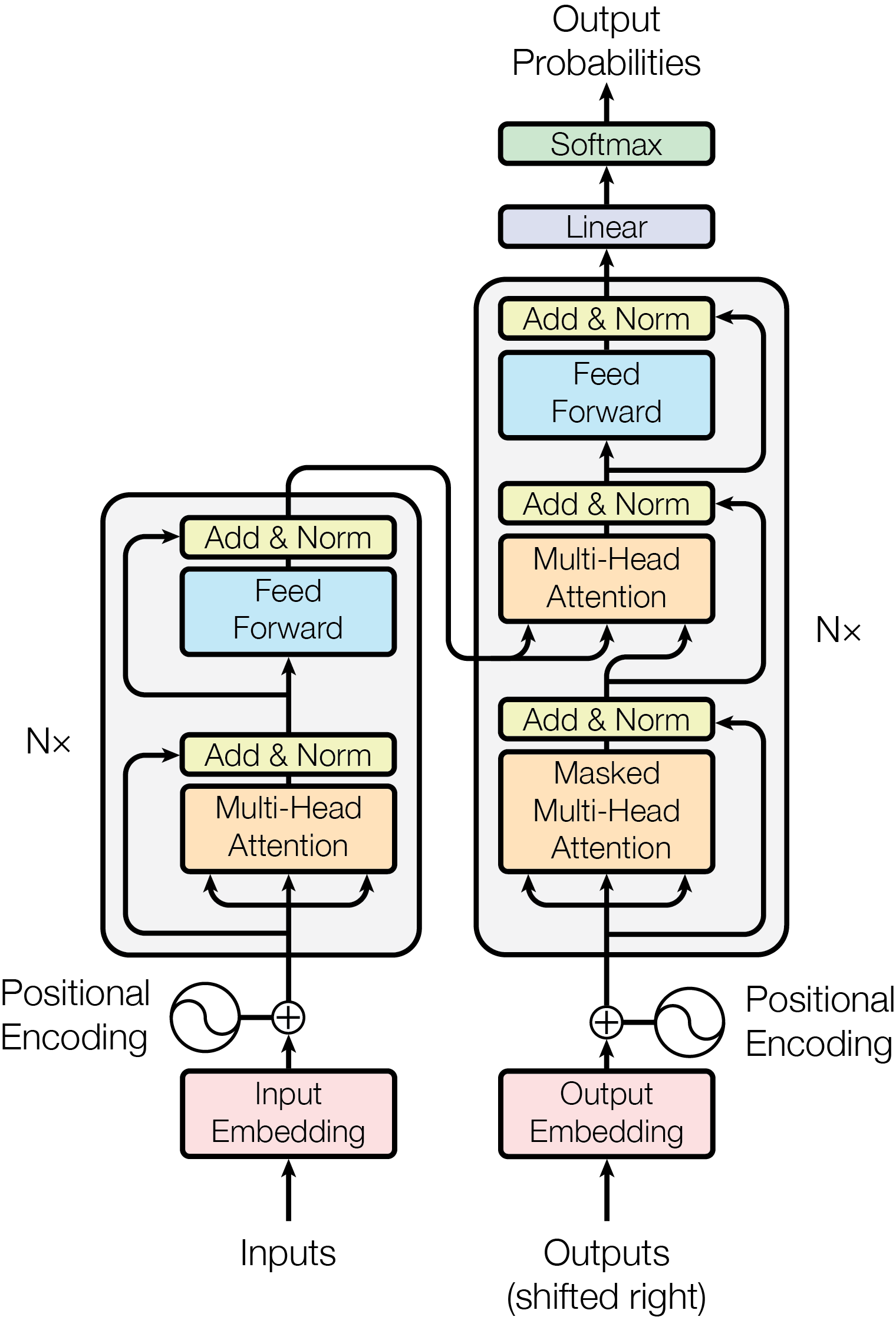}
    \vspace{.5em}
    \caption{The transformer model architecture \cite{vaswani2017attention}}
    \label{fig:transformer}
\end{figure}

The transformer \cite{vaswani2017attention}, as illustrated in Figure \ref{fig:transformer}, is an encoder-decoder model dispensing entirely with recurrence and convolutions. Instead, it leverages the attention mechanism to compute high-level contextualized embeddings. Being the first model to rely solely on attention mechanisms, it is able to address the issues commonly associated with recurrent neural networks, which factor computation along symbol positions of input and output sequences, and then precludes parallelization within samples. Despite this, the transformer is highly parallelizable and requires significantly less time to train. In the upcoming sections, we will highlight the recent breakthroughs in NLP involving transformer that changed the field overnight by introducing its designs, such as BERT \cite{devlin2018bert} and its improvements.

\subsection{BERT}
BERT \cite{devlin2018bert} is pre-trained using a combination of Masked Language Modeling (MLM) and Next Sentence Prediction (NSP) objectives. It provides high-level contextualized embeddings grasping the meaning of words in different contexts through global attention. As a result, the pre-trained BERT model can be fine-tuned for a wide range of downstream tasks, such as question answering and text classification, without substantial task-specific architecture modifications.

BERT and its variants allow the training of modern data-intensive models. Moreover, they are able to capture the contextual meaning of each piece of text in a way that traditional language models are unfit to do, while being quicker to develop and yielding better results with less data. On the other hand, BERT and other large neural language models are very expensive and computationally intensive to train/fine-tune and make inference.

\subsection{GPT-I, II, III}
GPT \cite{radford2018improving} is the first causal or autoregressive transformer-based model pre-trained using language modeling on a large corpus with long-range dependencies. However, its bigger an optimized version called GPT-2 \cite{radford2019language}, was pre-trained on WebText. Likewise, GPT-3 \cite{brown2020language} is architecturally similar to its predecessors. Its higher level of accuracy is attributed to its increased capacity and greater number of parameters, and it was pre-trained on Common Crawl. The OpenAI GPT family models has taken pre-trained language models by storm, they are very powerful on realistic human text generation and many other miscellaneous NLP tasks. Therefore, a small amount of input text can be used to generate large amount of high-quality text, while maintaining semantic and syntactic understanding of each word.

\subsection{ALBERT}
A lite BERT (ALBERT) \cite{lan2019albert} was proposed to address the problems associated with large models. It was specifically designed to provide contextualized natural language representations to improve the results on downstream tasks. However, increasing the model size to pre-train embeddings becomes harder due to memory limitations and longer training time. For this reason, this model arose.

ALBERT is a lighter version of BERT, in which next sentence prediction (NSP) is replaced by sentence order prediction (SOP). In addition to that, it employs two parameter-reduction techniques to reduce memory consumption and improve training time of BERT without hurting performance:
\begin{itemize}
  \item Splitting the embedding matrix into two smaller matrices to easily grow the hidden size with fewer parameters, ALBERT separates the hidden layers size from the size of the vocabulary embedding by decomposing the embedding matrix of the vocabulary.
  \item Repeating layers split among groups to prevent the parameter from growing with the depth of the network.
\end{itemize}

\subsection{RoBERTa}
The choice of language model hyper-parameters has a substantial impact on the final results. Hence, robustly optimized BERT pre-training approach (RoBERTa) \cite{liu2019roberta} is introduced to investigate the impact of many key hyper-parameters along with data size on model performance. RoBERTa is based on Google's BERT \cite{devlin2018bert} model and modifies key hyper-parameters, where the masked language modeling objective is dynamic and the NSP objective is removed. It is an improved version of BERT, pre-trained with much larger mini-batches and learning rates on a large corpus using self-supervised learning.

\subsection{XLNET}
The bidirectional property of transformer encoders, such as BERT \cite{devlin2018bert}, help them achieve better performance than autoregressive language modeling based approaches. Nevertheless, BERT ignores dependency between the positions masked, and suffers from a pretrain-finetune discrepancy when relying on corrupting the input with masks. In view of these pros and cons, XLNet \cite{yang2019xlnet} has been proposed. XLNet is a generalized autoregressive pretraining approach that allows learning bidirectional dependencies by maximizing the anticipated likelihood over all permutations of the factorization order. Furthermore, it overcomes the drawbacks of BERT \cite{devlin2018bert} due to its casual or autoregressive formulation, inspired from the transformer-XL \cite{dai2019transformer}.

\subsection{DistilBERT}
Unfortunately, the outstanding performance that comes with large-scale pretrained models is not cheap. In fact, operating them on edge devices under constrained computational training or inference budgets remains challenging. Against this backdrop, DistilBERT \cite{sanh2019distilbert} (or Distilled BERT) has seen the light to address the cited issues by leveraging knowledge distillation \cite{hinton2015distilling}.

DistilBERT is similar to BERT, but it is smaller, faster, and cheaper. It has 40\% less parameters than BERT base, runs 40\% faster, while preserving over 95\% of BERT's performance. It is trained using distillation of the pretrained BERT base model.

\subsection{XLM-RoBERTa}
Pre-trained multilingual models at scale, such as multilingual BERT (mBERT) \cite{devlin2018bert}  and cross-lingual language models (XLMs) \cite{lample2019cross}, have led to considerable performance improvements for a wide variety of cross-lingual transfer tasks, including question answering, sequence labeling, and classification. However, the multilingual version of RoBERTa \cite{liu2019roberta}  called XLM-RoBERTa \cite{conneau2019unsupervised}, pre-trained on the newly created 2.5TB multilingual CommonCrawl corpus containing 100 different languages, has further pushed the performance. It has shown strong improvements on low-resource languages compared to previous multilingual models.

\subsection{BART}
Bidirectional and auto-regressive transformer (BART) \cite{lewis2019bart} is a generalization of BERT \cite{devlin2018bert} and GPT \cite{radford2018improving}, it takes advantage of the standard transformer \cite{vaswani2017attention}. Concretely, it uses a bidirectional encoder and a left-to-right decoder. It is trained by corrupting text with an arbitrary noising function and learning a model to reconstruct the original text. BART has shown phenomenal success when fine-tuned on text generation tasks such as translation, but also performs well for comprehension tasks like question answering and classification.

\subsection{ConvBERT}
While BERT \cite{devlin2018bert} and its variants have recently achieved incredible performance gains in many NLP tasks compared to previous models, BERT suffers from large computation cost and memory footprint due to reliance on the global self-attention block. Although all its attention heads, BERT was found to be computationally redundant, since some heads simply need to learn local dependencies. Therefore, ConvBERT \cite{jiang2020convbert} is a better version of BERT \cite{devlin2018bert}, where self-attention blocks are replaced with new mixed ones that leverage convolutions to better model global and local context.

\subsection{Reformer}
Consistently, large transformer \cite{vaswani2017attention} models achieve state-of-the-art results in a large variety of linguistic tasks, but training them on long sequences is costly challenging. To address this issue, the Reformer \cite{kitaev2020reformer} was introduced to improve the efficiency of transformers while holding the high performance and the smooth training. Reformer is more efficient than transformer \cite{vaswani2017attention} thanks to locality-sensitive hashing attention and reversible residual layers instead of the standard residuals, and axial position encoding and other optimizations.

\subsection{T5}
Transfer learning has emerged as one of the most influential techniques in NLP. Its efficiency in transferring knowledge to downstream tasks through fine-tuning has given birth to a range of innovative approaches. One of these approaches is transfer learning with a unified text-to-text transformer (T5) \cite{raffel2020exploring}, which consists of a bidirectional encoder and a left-to-right decoder. This approach is reshaping the transfer learning landscape by leveraging the power of being pre-trained on a combination of unsupervised and supervised tasks and reframing every NLP task into text-to-text format.

\subsection{ELECTRA}
Masked language modeling (MLM) approaches like BERT \cite{devlin2018bert} have proven to be effective when transferred to downstream NLP tasks, although, they are expensive and require large amounts of compute. Efficiently learn an encoder that classifies token replacements accurately (ELECTRA) \cite{clark2020electra} is a new pre-training approach that aims to overcome these computation problems by training two Transformer models: the generator and the discriminator. ELECTRA trains on a replaced token detection objective, using the discriminator to identify which tokens were replaced by the generator in the sequences. Unlike MLM-based models, ELECTRA is defined over all input tokens rather than just a small subset that was masked, making it a more efficient pre-training approach.

\subsection{Longformer}
While previous transformers were focusing on making changes to the pre-training methods, the long-document transformer (Longformer) \cite{beltagy2020longformer} comes to change the transformer's self-attention mechanism. It has became the de facto standard for tackling a wide range of complex NLP tasks, with an new attention mechanism that scales linearly with sequence length, and then being able to easily process longer sequences. Longformer's new attention mechanism is a drop-in replacement for the standard self-attention and combines a local windowed attention with a task motivated global attention. Simply, it replaces the transformer \cite{vaswani2017attention} attention matrices with sparse matrices for higher training efficiency.

\subsection{DeBERTa}
DeBERTa \cite{he2020deberta} stands for decoding-enhanced BERT with disentangled attention. It is a pre-training approach that extends Google’s BERT \cite{devlin2018bert} and builds on the RoBERTa \cite{liu2019roberta}. Despite being trained on only half of the data used for RoBERTa, DeBERTa has been able to improve the efficiency of pre-trained models through the use of two novel techniques:
\begin{itemize}
  \item Disentangled attention (DA): an attention mechanism that computes the attention weights among words using disentangled matrices based on two vectors that encode the content and the relative position of each word respectively.
  \item Enhanced mask decoder (EMD): a pre-trained technique used to replace the output softmax layer. Thus, incorporate absolute positions in the decoding layer to predict masked tokens for model pre-training.
\end{itemize}

\section{Approach}
Transformer-based pre-trained language models have led to substantial performance gains, but careful comparison between different approaches is challenging. Therefore, we extend our study to uncover insights regarding their fine-tuning process and main characteristics. Our paper first aims to study the behavior of these models, following two approaches: a data-centric view focusing on the data state and quality, and a model-centric view giving more attention to the models tweaks. Indeed, we will see how data processing affects their performance and how adjustments and improvements made to the model over time is changing its performance. Thus, we seek to end with some takeaways regarding the optimal setup that aids in cross-validating a Transformer-based model, specifically model tuning hyper-parameters and data quality.

\subsection{Models Summary}
In this section, we present the base versions' details of the models introduced previously as shown in Table \ref{tab:timelineA1}.  We aim to provide a fair comparison based on the following criteria: L-Number of transformer layers, H-Hidden state size or model dimension, A-Number of attention heads, number of total parameters, tokenization algorithm, data used for pre-training, training devices and computational cost, training objectives, good performance tasks, and a short description regarding the model key points \cite{singh2021nlp}. All these information will help to understand the performance and behaviors of different transformer-based models and aid to make the appropriate choice depending on the task and resources.

\subsection{Configuration}
\begin{figure}[ht]
    \centering
    \includegraphics[scale=0.43]{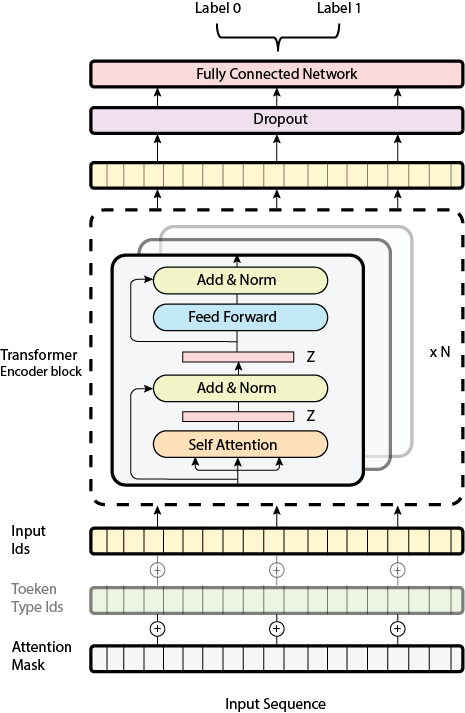}
    \vspace{.5em}
    \caption{The Architecture of the Transformer Encoder-Based Models.}
    \label{fig:encoder}
\end{figure}

\begin{figure}[ht]
    \centering
    \includegraphics[scale=0.47]{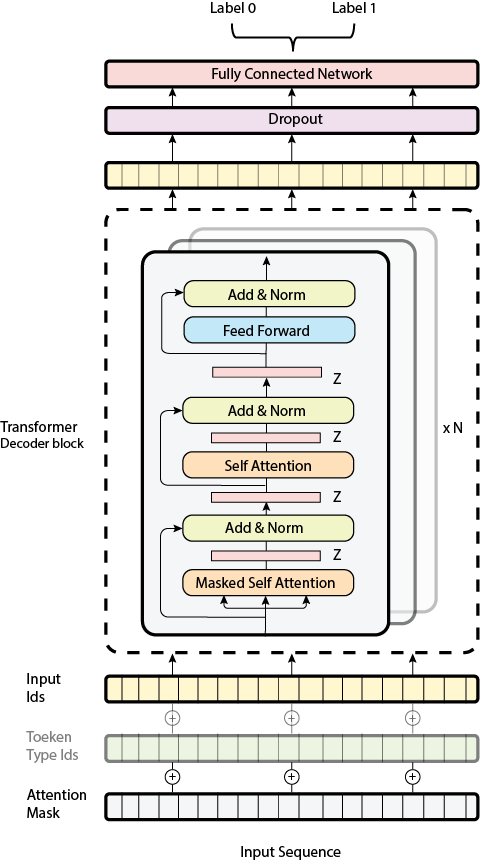}
    \vspace{.5em}
    \caption{The Architecture of the Transformer Decoder-Based Models.}
    \label{fig:decoder}
\end{figure}

It should be noted that we have used almost the same architecture building blocks for all our implemented models as shown in Figure \ref{fig:encoder} and Figure \ref{fig:decoder} for both encoder and decoder based models, respectively. In contrast, seq2seq models like BART are merely a bidirectional encoder pursued by an autoregressive decoder. Each model is fed with the three required inputs, namely input ids, token type ids, and attention mask. However, for some models, the position embeddings are optional and can sometimes be completely ignored (e.g RoBERTa), for this reason we have blurred them a bit in the figures. Furthermore, it is important to note that we uniformed the dataset in lower cases, and we tokenized it with tokenizers based on WordPiece \cite{wu2016google}, SentencePiece \cite{kudo2018sentencepiece}, and Byte-pair-encoding \cite{sennrich2015neural} algorithms.

In our experiments, we used a highly optimized setup using only the base version of each pre-trained language model. For training and validation, we set a batch size of 8 and 4, respectively, and fine-tuned the models for 4 epochs over the data with maximum sequence length of 384 for the intent of correspondence to the majority of reviews' lengths and computational capabilities. The AdamW optimizer is utilized to optimize the models with a learning rate of 3e-5 and the epsilon (eps) used to improve numerical stability is set to 1e-6, which is the default value. Furthermore, the weight decay is set to 0.001, while excluding bias, LayerNorm.bias, and LayerNorm.weight from the decay weight when fine-tuning, and not decaying them when it is set to 0.000. We implemented all of our models using PyTorch and transformers library from Hugging Face, and ran them on an NVIDIA Tesla P100-PCIE GPU-Persistence-M (51G) GPU RAM.

\subsection{Evaluation}
\begin{figure}[!ht]
\centering
  \includegraphics[scale=0.49]{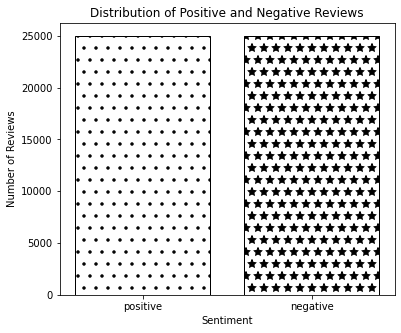}
      \vspace{.3em}
 \caption{Positive and negative reviews distribution}
  \label{fig:reviewsdist}
\end{figure}

\textbf{Dataset} to fine-tune our models, we used the IMDb movie review dataset \cite{maas-EtAl:2011:ACL-HLT2011}. A binary sentiment classification dataset having 50K highly polar movie reviews labelled in a balanced way between positive and negative. We chose it for our study because it is often used in research studies and is a very popular resource for researchers working on NLP and ML tasks, particularly those related to sentiment analysis and text classification due to its accessibility, size, balance and pre-processing. In other words, it is easily accessible and widely available, with over 50K reviews well-balanced, with an equal number of positive and negative reviews as shown in Figure \ref{fig:reviewsdist}. This helps prevent biases in the trained model. Additionally, it has already been pre-processed with the text of each review cleaned and normalized.

\textbf{Metrics} To assess the performance of the fine-tuned transformers on the IMDb movie reviews dataset, tracking the loss and accuracy learning curves for each model is an effective method. These curves can help detect incorrect predictions and potential overfitting, which are crucial factors to consider in the evaluation process. Moreover, widely-used metrics, namely accuracy, recall, precision, and F1-score are valuable to consider when dealing with classification problems. These metrics can be defined as:

\begin{equation}
   Precision = \frac{TP}{TP+FP}
   \quad\text{,}
   \quad
   Recall = \frac{TP}{TP+FN}
   \quad\text{, and}
   \quad
   F1= 2\times\frac{Precision \times Recall}{Precision + Recall}
\end{equation}

\section{Results}
\begin{figure}[ht]
  \centering
    \includegraphics[scale=0.40]{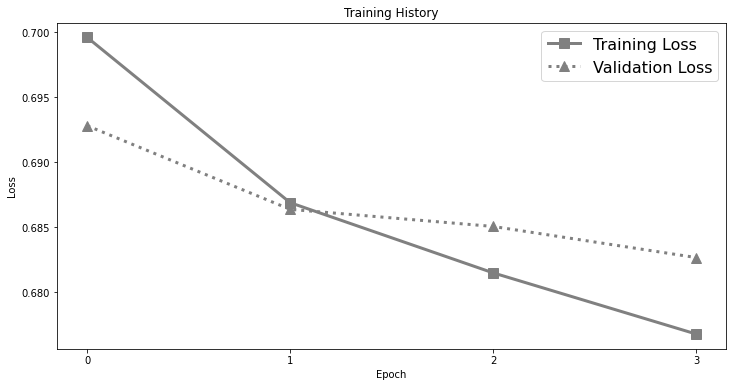}
    \vspace{.7em}
    \caption{Worst model: GPT-2 loss learning curve}
    \label{fig:lossworstmodel}
\end{figure}

\begin{figure}[ht]
  \centering
    \includegraphics[scale=0.40]{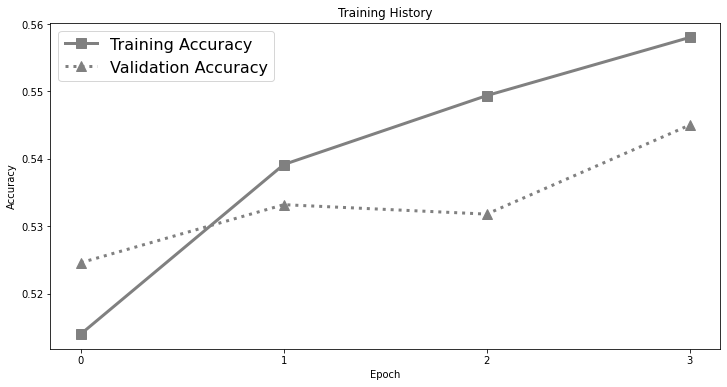}
        \vspace{.5em}
    \caption{Worst model: GPT-2 acc learning curve}
    \label{fig:accworstmodel}
\end{figure}

In this section, we present the fine-tuning main results of our implemented transformer-based language models on the opinion mining task on the IMDb movie reviews dataset. Typically, all the fine-tuned models perform well with fairly high performance, except the three autoregressive models: GPT, GPT-2, and Reformer, as shown in Table \ref{tab:mainresults}. The best model, ELECTRA, provides an F1-score of 95.6 points, followed by RoBERTa, Longformer, and DeBERTa, with an F1-score of 95.3, 95.1, and 95.1 points, respectively. On the other hand, the worst model, GPT-2 provide an F1-score of 52.9 points as shown in Figure \ref{fig:lossworstmodel} and \linebreak Figure  \ref{fig:accworstmodel}. From the results, it is clear that purely autoregressive models do not perform well on comprehension tasks like sentiment classification, where sequences may require access to bidirectional contexts for better word representation, therefore, good classification accuracy. Whereas, with autoencoding models taking advantage of left and right contexts, we saw good performance gains. For instance, the autoregressive XLNet model is our fourth best model in Table \ref{tab:mainresults} with an F1 score of 94.9\%, it incorporates modelling techniques from autoencoding models into autoregressive models while avoiding and addressing limitations of encoders. The code and fine-tuned models are available at \cite{opinion2023zekaoui}.

\definecolor{top1}{RGB}{255, 0, 255}
\definecolor{top2}{RGB}{255, 102, 255}
\definecolor{top3}{RGB}{255, 153, 255}
\definecolor{top4}{RGB}{255, 204, 255}

\begin{table*}[]
    \centering
    \fontsize{8pt}{10pt}\selectfont
     \caption {Transformer-based language models validation performance on the opinion mining IMDb dataset}
     \label{tab:mainresults}
    \begin{tabular}{lccccc}
        \hline
            Model & Recall & Precision & F1 & Accuracy  \\
        \hline
            BERT             &  93.9    &  94.3    &  94.1    &  94.0    \\
            GPT              &  92.4    &  51.8    &  66.4    &  53.2    \\
            GPT-2            &  51.1    &  54.8    &  52.9    &  54.5    \\
            ALBERT           &  94.1    &  91.9    &  93.0    &  93.0    \\
            RoBERTa          &  96.0    &  94.6    &  \colorbox{top2}{95.3}    &  \colorbox{top2}{95.3}    \\
            XLNet            &  94.7    &  95.1    &  \colorbox{top4}{94.9}    &  94.8    \\
            DistilBERT       &  94.3    &  92.7    &  93.5    &  93.4    \\
            XLM-RoBERTA      &  83.1    &  71.7    &  77.0    &  75.2    \\
            BART             &  96.0    &  93.3    &  94.6    &  94.6    \\
            ConvBERT         &  95.5    &  93.7    &  94.6    &  94.5    \\
            DeBERTa          &  95.2    &  95.0    &  \colorbox{top3}{95.1} &  \colorbox{top3}{95.1}  \\
            \textbf{ELECTRA} &  95.8    &  95.4    &  \colorbox{top1}{95.6} &  \colorbox{top1}{95.6} \\
            Longformer       &  95.9    &  94.3    &  \colorbox{top3}{95.1} &  \colorbox{top4}{95.0}    \\
            Reformer         &  54.6    &  52.1    &  53.3    &  52.2    \\
            T5               &  94.8    &  93.4    &  94.0    &  93.9    \\
        \hline
    \end{tabular}  
\end{table*}


\section{Ablation Study}
\begin{figure}[ht]
	\centering
	\includegraphics[scale=0.40]{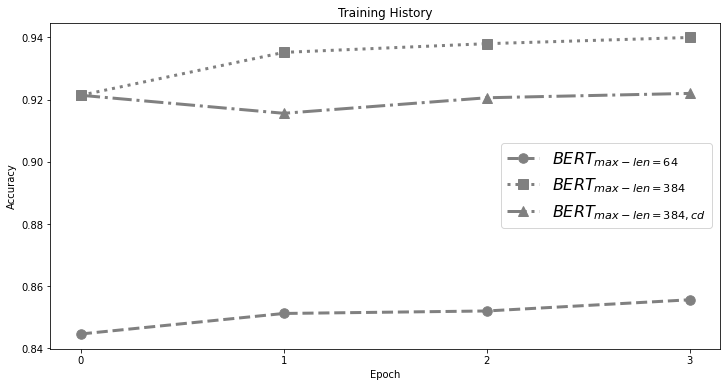}
        \vspace{.7em}
    \caption{Validation accuracy history of BERT model based on different configurations}
    \label{fig:ablaacc}
\end{figure}

 In Table \ref{tab:ablation} and Figure \ref{fig:ablaacc}, we demonstrate the importance of configuration choices through controlled trials and ablation experiments. Indeed, the maximum length of the sequence and data cleaning are particularly crucial. Thus, to make our ablation study credible, we fine-tuned our BERT model with the same setup, changing only the sequence length (max-len) initially and cleaning the data (cd) at another time to observe how they affect the performance of the model.

\begin{table}[ht]
    \centering
    \fontsize{8pt}{10pt}\selectfont
    \caption {Validation results of the BERT model based on different configurations, where cd stands for cleaned data, meaning that the latest model (BERT\textsubscript{max-len=384, cd}) is trained on an exhaustively cleaned text}
    \begin{tabular}{lcccc}
        \hline
            Model & Recall & Precision & F1 & Accuracy\\
        \hline
            BERT\textsubscript{max-len=64}       &  86.8\%     &  84.7\%     &  85.8\%     &  85.6\%   \\
            BERT\textsubscript{max-len=384}      &  \textbf{93.9\%}  &  \textbf{94.3\%}  &  \textbf{94.1\%}  &  \textbf{94.0\%} \\
            BERT\textsubscript{max-len=384, cd}  &  92.6\%     &  91.6\%     &  92.1\%     &  92.2\%   \\
        \hline
    \end{tabular}
\label{tab:ablation}
\end{table}

\subsection{Effects of Hyper-Parameters}
\begin{figure}[ht]
    \centering
    \includegraphics[scale=0.38]{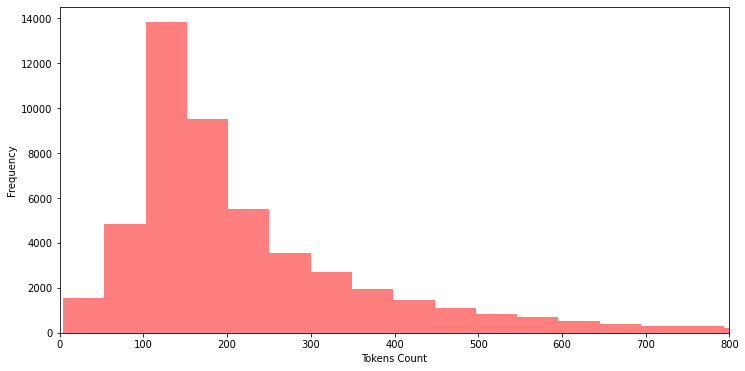}
        \vspace{.5em}
    \caption{Distribution of the number of tokens for a better selection of the maximum sequence length}
    \label{fig:tkncount}
\end{figure}

The gap between the performance of BERT\textsubscript{max-len=64} and BERT\textsubscript{max-len=384} on the IMDb dataset is an astounding 8.3 F1 points, as in Table \ref{tab:ablation}, demonstrating how important this parameter is. Thereby, visualizing the distribution of tokens or words count is the ultimate solution for defining the optimal and correct value of the maximum length parameter that corresponds to all the training data points. Figure \ref{fig:tkncount} illustrates the distribution of the number of tokens in the IMDb movie reviews dataset, it shows that the majority of reviews are between 100 and 400 tokens in length. In this context, we chose 384 as the maximum length reference to study the effect of the maximum length parameter, because it covers the majority of review lengths while conserving memory and saving computational resources. It should be noted that the BERT model can process texts up to 512 tokens in length. It is a consequence of the model architecture and can not be adjusted directly.

\subsection{Effects of Data Cleaning} 
Traditional machine learning algorithms require extensive data cleaning before vectorizing the input sequence and then feeding it to the model, with the aim of improving both reliability and quality of the data. Therefore, the model can only focus on important features during training. Contrarily, the performance dropped down dramatically by 2 F1 points when we cleaned the data for the BERT model. Indeed, the cleaning carried out aims to normalize the words of each review. It includes lemmatization to group together the different forms of the same word, stemming to reduce a word to its root, which is affixed to suffixes and prefixes, deletion of URLs, punctuations, and patterns that do not contribute to the sentiment, as well as the elimination of all stop words, except the words ``no", ``nor", and ``not", because their contribution to the sentiment can be tricky. For instance, ``Black Panther is boring" is a negative review, but ``Black Panther is not boring" is a positive review. This drop can be justified by the fact that BERT model and attention-based models need all the sequence words to better capture the meaning of words' contexts. However, with cleaning, words may be represented differently from their meaning in the original sequence. Note that ``not boring" and ``boring" are completely different in meaning, but if the stop word ``not" is removed, we end up with two similar sequences, which is not good in sentiment analysis context.

\subsection{Effects of Bias and Training Data} 
\begin{figure}[ht]
	\centering
	\includegraphics[scale=0.38]{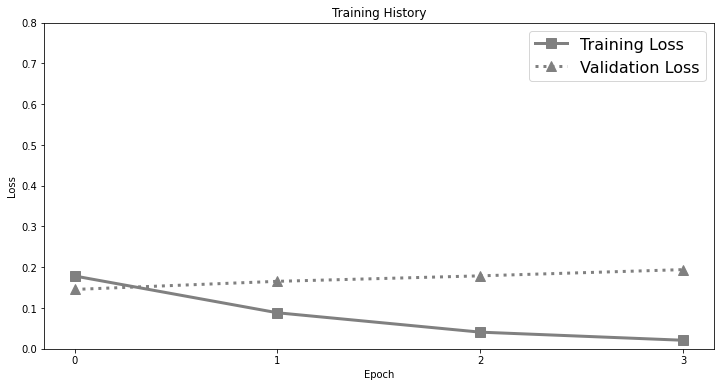}
    \vspace{.5em}
    \caption{Best model: ELECTRA loss learning curve}
    \label{fig:lossbestmodel}
  \end{figure}

\begin{figure}[ht]
	\centering
	\includegraphics[scale=0.38]{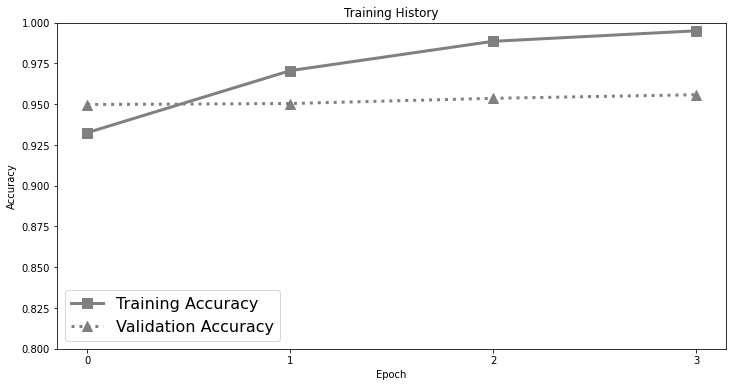}
        \vspace{.5em}
    \caption{Best model: ELECTRA acc. learning curve}
    \label{fig:accbestmodel}
\end{figure}

Carefully observing the accuracy and the loss learning curves in Figure \ref{fig:lossbestmodel} and Figure \ref{fig:accbestmodel}, we notice that the validation loss starts to creep upward and the validation accuracy starts to go down. In this perspective, the model in question continues to lose its ability to generalize well on unseen data. In fact, the model is relatively biased due to the effect of the training data and data-drift issues related to the fine-tuning data. In this context, we assume that the model starts to overfit. However, setting different dropouts, reducing the learning rate, or even trying larger batches will not work. On the other hand, these strategies sometimes give worst results, then a more critical overfitting problem. For this reason, pretraining these models on your industry data and vocabulary and then fine-tuning them may be the best solution.

\section{Conclusion}
In this paper, we presented a detailed comparison to highlight the main characteristics of transformer-based pre-trained language models and what differentiates them from each other. Then, we studied their performance on the opinion mining task. Thereby,  we deduce the power of fine-tuning and how it helps in leveraging the pre-trained models' knowledge to achieve high accuracy on downstream tasks, even with the bias they came with due to the pre-training data. Experimental results show how performant these models are. We have seen the highest F1-score with the ELECTRA model with 95.6 points, across the IMDb dataset. Similarly, we found that access to both left and right contexts is necessary when it comes to comprehension tasks like sentiment classification. We have seen that autoregressive models like GPT, GPT-2, and Reformer perform poorly and fail to achieve high accuracy. Nevertheless, XLNet has reached good results even though it is an autoregressive model because it incorporates ideas taken from encoders characterized by their bidirectional property. Indeed, all performances were nearby, including DistilBERT, which helps to gain incredible performance in less training time thanks to knowledge distillation. For example, for 4 epochs, BERT took 70 minutes to train, while DistilBERT took 35 minutes, losing only 0.6 F1 points, but saving half the time taken by BERT. Moreover, our ablation study shows that the maximum length of the sequence is one of the parameters having a significant impact on the final results and must be carefully analyzed and adjusted. Likewise, data quality is a must for good performance, data that will do not need to be processed, since extensive data cleaning processes may not help the model capture local and global contexts in sequences, distilled sometimes with words removed or trimmed during cleaning. Besides, we notice, that the majority of the models we fine-tuned on the IMDb dataset start to overfit at a certain number of epochs, which can lead to biased models. However, good quality data is not even enough, but pre-training a model on large amounts of business problem data and vocabulary may help on preventing it from making wrong predictions and may help on reaching a high level of generalization.

\section*{Acknowledgments}
We are grateful to the Hugging Face team for their role in democratizing state-of-the-art machine learning and natural language processing technology through open-source tools. Their commitment to providing valuable resources to the research community is highly appreciated, and we acknowledge their vital contribution to the development of our article.

\section*{Appendix}
Appendix for ``Analysis of the Evolution of Advanced Transformer-based Language Models: Experiments on Opining Mining"

\setcounter{table}{0}
\renewcommand{\thetable}{A\arabic{table}}

\begin{center}
\small
\newgeometry{vmargin={25mm}, hmargin={15mm,15mm}}
\begin{landscape}
\footnotesize
\begin{longtable}{ | p{1.3cm} | p{0.3cm} | p{0.7cm} | p{0.3cm}| p{0.8cm} | p{0.8cm} |p{1.6cm} | p{2.5cm} | p{1.9cm} | p{1.9cm} | p{2.8cm} | p{3cm}|} 
\caption{Summary and comparison of transformer-based models.}
\label{tab:timelineA1}
\\ 
\hline
\centering

\textbf{Model} & 
\textbf{L}  & 
\textbf{H}  & 
\textbf{A}  & 
\textbf{Att. Type}  & 
\textbf{Total Params}  & 
\textbf{Tokenization}  & 
\textbf{Training Data}  & 
\textbf{Computational Cost}  &
\textbf{Training Objectives}  & 
\textbf{Performance Tasks}  & 
\textbf{Short Description}  \\ 
\hline
    GPT
    & 12 
    & 512
    & 12 
    & Global 
    & 110M 
    & Byte-Pair-Encoding \cite{sennrich2015neural}
    & Books Corpus (800M words)
    & -
    & Autoregressive, decoder
    & Zero-shot, Text Summarization, question answering, translation.
    & The first Transformer-based autoregressive and Causal Masking model.
    
    \\
    
    BERT         
    & 12 
    & 768 
    & 12 
    & Global 
    & 110M 
    & WordPiece \cite{wu2016google}
    & Books Corpus (800M words) and English Wikipedia (2,500M words) 
    & 4 days on 4 Cloud TPUs in Pod configuration.
    & Autoencoding, Encoder (MLM - NSP)
    & Text classification, Natural Language Inference, Question Answering.
    & The first Transformer-based autoencoding model, that uses global attention to provide high-level bidirectional contextualization.
    
    \\
    
    GPT-2       
    & 12
    & 1600
    & 12 
    & Global 
    & 117M
    & Byte-Pair-Encoding 
    & WebText (10B words) 
    & - 
    & Autoregressive, Decoder
    & Zero-shot, Text Summarization, question answering, translation.
    & Optimized and bigger than GPT and performs well on zero-shot settings.
    
    \\
    
    GPT-3        
    & 96 
    & 12288
    & 96 
    & Global 
    & 175B 
    & Byte-Pair-Encoding  
    & Filtered Common Crawl, WebText2, Books1, Books2, and Wikipedia for 300B words.
    & -
    & Autoregressive, Decoder
    & Text Summarization, Question Answering, Translation, Zero-shot, One-shot, Few-shot.
    & Bigger that its predecessors.
    
    \\
    
    ALBERT       
    & 12 
    & 768
    & 12 
    & Global 
    & 11M 
    & SentencePiece \cite{kudo2018sentencepiece} 
    & Books Corpus \cite{zhu2015aligning} and English Wikipedia.
    & Cloud TPU V3 TPUs number ranges from 64 to 512 (32h ALBERT-xxlarge).
    & Autoencoding, Encoder, Sentence-Ordering Prediction (SOP)
    & Semantic Similarity, Semantic Relevance, Question Answering, Reading Comprehension.
    & Smaller and similar to BERT with minimal tweaks including the splitting of layers into groups via cross-layer parameter sharing, making it faster and reducing memory footprint.
    
    \\
    
    DistilBERT   
    & 6 
    & 768
    & 12
    & Global 
    & 66M 
    & WordPiece 
    & English Wikipedia and Toront Book Corpus.
    & 90 hours on 8 16GB V100 GPUs.
    & Autoencoding (MLM), Encoder
    & Semantic Similarity, Semantic Relevance, Question Answering, Textual Entailment.
    & Pre-training leveraging knowledge distillation to deliver great results as BERT with lower latency.  Similar to BERT model but smaller.
    
    \\
    
    RoBERTa      
    & 12 
    & 1024
    & 12
    & Global
    & 125M 
    & Byte-Pair-Encoding 
    & Book Corpus \cite{zhu2015aligning}, CC-News, Open Web Text, and Stories \cite{trinh2018simple}.
    & 8 × 32GB Nvidia V100 GPUs.
    & Autoencoding (Dynamic MLM, No NSP), Encoder
    & Text Classification, Language Inference, Question Answering.
    & Pre-trained with large batches using some tricks for diverse learning like dynamic masking, where tokens are differently masked for each epoch.
    
    \\
    
    XLM      
    & 12 
    & 2048
    & 8
    & Global
    & - 
    & Byte-Pair Encoding 
    & Wikipedias of the XNLI languages.
    & 64 Volta GPUs for the language modeling tasks and 8 GPUs for the MT tasks.
    & Autoencoding, Encoder, Causal Language Modeling (CLM), Masked Language Modeling (MLM), and Translation Language Modeling (TLM).
    & Translation tasks and NLU cross-lingual benchmarks.
    & By being trained on several pre-training objectives on a multilingual corpus, XLM proves that multilingual pre-training methods have a strong impact, especially on the performance of multilingual tasks.
    
    \\
    
    XLM-RoBERTa      
    & 12 
    & 768
    & 8
    & Global
    & 270M 
    & SentencePiece 
    & CommonCrawl Corpus in 100 languages.
    & 100 32GB Nvidia V100 GPUs.
    & Autoencoding, Encoder, Masked Language Modeling (MLM).
    & Translation tasks and NLU cross-lingual benchmarks.
    & Using only the masked language modeling objective, XLM-RoBERTa uses RoBERTa tricks on XLM approaches. it is able to detect the input language by itself (100 languages).
    
    \\

    ELECTRA      
    & 12 
    & 768
    & 12
    & Global
    & 110M 
    & WordPiece 
    & Wikipedia, BooksCorpus, Gigas5 \cite{parker2011english}, ClueWeb 2012-B, and Common Crawl.
    & 4 days on 1 GPU.
    & Generator (autoregressive, Replaced Token Detection) and Discriminator (Electra: Predicting Masked Tokens).
    & Sentiment Analysis, Language Inference Tasks.
    & Replaced token detection is a pre-training objective that addresses MLM issues and it resutls in efficient performance.
    
    \\
    
    DeBERTa      
    & 12 
    & 768
    & 12
    & Global (Disentangled attention)
    & ~125M 
    & Byte-Pair Encoding 
    & Wikipedia, BooksCorpus, Reddit content, Stories, STORIES.
    & 10 days 64 V100 GPUs.
    & Autoencoding, Disentangled Attention Mechanism, and Enhanced Mask Decoder.
    & DeBERTa was the first pretrained model to beat HLP on the SuperGLUE benchmark \cite{wang2019superglue}.
    & DeBERTa uses RoBERTa with Disentangled attention and an enhanced mask decoder to significantly improve model performance on many downstream tasks while being trained only on half of the data used in RoBERTa large version.
    
    \\
    
    XLNet      
    & 12 
    & 768
    & 12
    & Global
    & 110M 
    & SentencePiece 
    & Wikipedia, BooksCorpus, Gigas5 \cite{parker2011english}, ClueWeb 2012-B, and Common Crawl.
    & 5.5 days on 512 TPU v3 chips.
    & Autoregressive, Decoder
    & XLNet achieved state-of-the-art results and outperformed BERT on 20 downstream task inlcuding sentiment analysis, question answering, Reading Comprehension, Document Ranking.
    & XLNet incorporates ideas from Transformer-XL \cite{dai2019transformer} and addresses the pretrain-finetune BERT's discrepancy being more capable to grasp dependencies between masked tokens.
    
    \\

    BART      
    & 12 
    & 768
    & 16
    & Global
    & 139M
    & Byte-Pair Encoding 
    & Wikipedia, BooksCorpus.
    & -
    & Generative sequence to sequence, Encoder Decoder, Token Masking, Token Deletion, Text Infilling, Sentence Permutation, and Document Rotation.
    & BART beats its predecessors on generation tasks such as translation and achieved state-of-the-art results, while performing similarly to RoBERTa on discriminative tasks including question answering and classification.
    & Trained to map corrupted text to the original using an arbitrary noising function.
    
    \\

    ConvBERT      
    & 12
    & 768
    & 12
    & Global
    & 124M 
    & WordPiece 
    & OpenWebText \cite{gokaslan2019openwebtext}
    & GPU and TPU
    & Autoencoding, Encoder
    & With fewer parameters and lower costs ConvBERT consistently outperforms BERT on various downstream tasks with less training cost.
    & For reduced redundancy and better modeling of global and local context, BERT's self-attention blocks are replaced by mixed-attention blocks incorporating self-attention and span-based dynamic convolutions.
    
    \\
    
    Reformer      
    & 12
    & 1024
    & 8
    & Attenion with Local Sensitive Hashing
    & 149M
    & SentencePiece 
    & OpenWebText \cite{gokaslan2019openwebtext}
    & Parallelization across 8 GPUs or 8 TPU v3 cores.
    & Autoregressive, Decoder.
    & Performs well with paragmatic requirements, thanks to reduction of the attention complexity.
    & An efficient and faster Transformer that costs less time on long sequences thanks to two optimization techniques, Local-Sensitive Hashing attention and axial position encoding.
    
    \\

    T5      
    & 12
    & 768
    & 12
    & Global
    & ~220M
    & SentencePiece
    & The Colossal Clean Crawled Corpus (C4)
    & Cloud TPU Pods.
    & Generative sequence to sequence, Encoder-Decoder.
    & Entailement, Coreference challenges, Question Answering Tasks via SuperGLUE benchmark
    & To incorporate the varieties of most linguistic tasks, T5 pre-trained on a mix of supervised and unsupervised tasks in a text-to-text format. 
    
    \\
    
    Longformer      
    & 12
    & 768
    & 12
    & \begin{center}Local + Global.\end{center}
    & ~149M
    & Byte-Pair-Encoding
    & Books corpus, English Wikipedia, and Realnews dataset \cite{zellers2019defending}
    & .
    & Autoregressive, Decoder
    & Longformer achieved state-of-the-art results on two benchmark datasets WikiHop and TriviaQA.
    & For heigher training efficiency on long documents, Longformer uses sparse matrices instead of attention matrices to linearly scale with sequences of length up to 4 096.
    
    \\
    
    \hline
\end{longtable}
\end{landscape}
\end{center}

\bibliographystyle{unsrt}
\small
\bibliography{main}  






\end{document}